\documentclass{article}
\newif\ifanonymous \anonymousfalse   
\usepackage{xcolor}
\ifanonymous\usepackage{corl_2026}\else\usepackage[final]{corl_2026}\fi
\usepackage{amsmath,amssymb}
\usepackage{graphicx}
\usepackage{booktabs}
\usepackage{microtype}
\usepackage{float}
\usepackage{tikz}
\usepackage{enumitem}
\hypersetup{hypertexnames=false,pdfsubject={CoRL 2026 Workshop: Efficient Foundation Models for Real-Time Embodied AI}}
\newcommand{\R}{\mathbb{R}}
\makeatletter\renewcommand{\@noticestring}{Workshop on Efficient Foundation Models for Real-Time Embodied AI, Conference on Robot Learning (CoRL 2026), Austin, USA.}\makeatother


\title{Staleness Deadlines for Motion Plans during Onboard Reasoning}
\author{Hassan Iqbal\\
The University of Texas at Austin\\
\texttt{hassan.iqbal@utexas.edu}}

\begin{document}
\maketitle

\begin{abstract}
Language models may enable general autonomy in robot planning, but their reasoning on an onboard computer can run for most of a minute while the robot and nearby pedestrians keep moving.
The robot must then decide how long to keep following a plan that the language model's answer may overturn.
Here we present \emph{staleness deadlines}, which hand that decision to the robot's clock.
Calibrated prediction regions for pedestrians and deviation tubes for the robot set the deadline.
When the language model answers, a graph planner rebuilds the route, and the robot switches to it only after the new plan passes the same clearance check against walls and predicted pedestrians.
We use exact binomial tests to determine which prespecified budget-selection policies keep close calls, wrong decisions and late arrivals below their targets at each inference profile.
Candidate policies either fix one budget or choose a budget for each query from how long the robot can continue its current plan before it must brake.
We study corridor signs that close routes and delivery requests whose destination the language model must infer.
In 81,000 simulated missions with recorded Qwen3-4B answers, full reasoning cuts wrong decisions about twentyfold, and tracking until the deadline cuts late arrivals by up to 64\%.
The robot never collides while it moves, although pedestrians who never yield still walk into the robot once it has stopped, in 23\% of offline and 32\% of live missions.
On a Jetson AGX Orin under perception load, short budgets fail the wrong-decision target and full reasoning fails the lateness target.
Between these extremes, a fixed-budget policy passes for corridor signs and prespecified per-query policies pass for deliveries.
Live Orin missions agree with offline replay on all three losses in 24 of the 32 missions where either run records any of them.
The deadline turns slower reasoning into measurable mission outcomes, which let designers test how much reasoning can be supported under each inference profile.
\end{abstract}


\section{Introduction}
\begin{figure}[t]
\centering
\scalebox{0.74}{
\begin{tikzpicture}[font=\scriptsize, >=stealth]
  \definecolor{planblue}{RGB}{38,98,170}
  \definecolor{tubeblue}{RGB}{186,208,236}
  \definecolor{personor}{RGB}{214,110,30}
  \definecolor{brakered}{RGB}{196,52,52}
  \definecolor{thinkpurple}{RGB}{110,70,170}
  \definecolor{walkgreen}{RGB}{43,122,75}
  \definecolor{newteal}{RGB}{20,128,140}
  \definecolor{checkgold}{RGB}{226,170,40}
  \definecolor{signyellow}{RGB}{250,241,205}
  \node[anchor=north west, font=\small\bfseries] at (-0.05,4.62) {(a)};
  \node[anchor=north west, draw=signyellow!45!black, fill=signyellow, rounded corners=2pt, text width=5.95cm,
        inner sep=2.5pt, align=left] (sign) at (0.5,4.6)
    {\textbf{Sign.} Maintenance closes the east corridor at 1:15 pm for 60 minutes. It is 1:43 pm.};
  \node[anchor=north west, draw=thinkpurple!55, fill=thinkpurple!6, rounded corners=2pt, text width=5.95cm,
        inner sep=2.5pt, align=left] (think) at ([yshift=-0.2cm]sign.south west)
    {\textbf{\textcolor{thinkpurple}{The language model thinks.}} \textcolor{black!70}{\emph{1:15 pm plus 60 minutes is 2:15 pm, so at 1:43 pm the east corridor is still closed\ldots}}};
  \node[anchor=north west, draw=newteal!70, fill=newteal!8, rounded corners=2pt, text width=5.95cm,
        inner sep=2.5pt, align=left] (answer) at ([yshift=-0.2cm]think.south west)
    {\textbf{\textcolor{newteal}{Answer.}} east corridor: closed, north corridor: open};
  \draw[->, gray!70] ([xshift=-2.3cm]sign.south) -- ([xshift=-2.3cm]think.north);
  \draw[->, gray!70] ([xshift=-2.3cm]think.south) -- ([xshift=-2.3cm]answer.north);
  \node[anchor=north west, font=\small\bfseries] at (6.6,4.62) {(b)};
  \node[anchor=north west, draw=planblue!45, fill=planblue!6, rounded corners=2pt, text width=4.35cm,
        inner sep=2.5pt, align=left, execute at begin node={\hyphenpenalty=10000\exhyphenpenalty=10000}] (nurse) at (7.18,4.6)
    {\textbf{Nurse.} Please take these 4 swab samples to a lab\ldots{} A power test closes the east lab at 8:15 am today for 120 minutes. The north lab scans each tube in 5 seconds\ldots{} It is Monday at 8:31 am.};
  \node[anchor=north west, inner sep=1pt, text=thinkpurple] (nthink) at ([xshift=0.05cm, yshift=-0.1cm]nurse.south west)
    {\emph{thinks for a median 54 s on the Orin}};
  \node[anchor=north west, draw=newteal!70, fill=newteal!8, rounded corners=2pt, text width=4.35cm,
        inner sep=2.5pt, align=left] (nanswer) at ([xshift=-0.05cm, yshift=-0.1cm]nthink.south west)
    {\textbf{\textcolor{newteal}{Answer.}} lab: north, handoff: 20};
  \begin{scope}[xshift=12.05cm, yshift=2.25cm, scale=0.86]
    \node[anchor=north west, font=\small\bfseries] at (-0.2,2.45) {(c)};
    \fill[gray!8] (0,0) rectangle (5.0,1.1);
    \fill[gray!8] (2.7,1.1) rectangle (3.8,2.05);
    \draw[gray!65, line width=1.2pt] (0,0) -- (5.0,0);
    \draw[gray!65, line width=1.2pt] (0,1.1) -- (2.7,1.1) -- (2.7,2.05);
    \draw[gray!65, line width=1.2pt] (3.8,2.05) -- (3.8,1.1) -- (5.0,1.1);
    \node[gray!80!black, anchor=south east] at (4.95,0.04) {east corridor};
    \node[gray!80!black, rotate=90, anchor=east] at (2.55,2.03) {north};
    \fill[tubeblue, opacity=0.85] (0.45,0.48) -- (2.25,0.34) -- (2.25,0.76) -- (0.45,0.62) -- cycle;
    \draw[planblue, dashed, line width=0.7pt, ->] (0.45,0.55) -- (4.85,0.55);
    \node[planblue, anchor=south west] at (0.25,0.7) {old plan $p_0$};
    \node[planblue!80!black, anchor=north west] at (0.3,0.33) {deviation tube};
    \draw[brakered, line width=1.4pt] (2.25,0.32) -- (2.25,0.78);
    \node[brakered, anchor=south, align=center] at (1.5,1.15) {stops here if it\\brakes at $T_{\mathrm{safe}}$};
    \draw[brakered, ->, line width=0.4pt] (1.85,1.17) -- (2.2,0.84);
    \fill[planblue] (0.45,0.55) circle (0.12);
    \fill[personor!45, opacity=0.55] (3.25,0.58) circle (0.4);
    \fill[personor!45, opacity=0.55] (3.25,1.15) circle (0.27);
    \fill[personor!45, opacity=0.55] (3.25,1.58) circle (0.18);
    \fill[personor] (3.25,1.9) circle (0.09);
    \node[personor!85!black, anchor=west, align=left] at (3.9,1.62) {person and\\prediction\\regions};
    \node[font=\small\bfseries, thinkpurple] at (2.55,0.55) {?};
  \end{scope}
  \begin{scope}[xshift=2.35cm, yshift=0cm, x=0.47cm]
    \node[anchor=north west, font=\small\bfseries] at (-5.3,1.98) {(d)};
    \node[anchor=east] at (-0.4,1.32) {language model};
    \node[anchor=east] at (-0.4,0.66) {robot};
    \draw[brakered, dashed, line width=0.6pt] (7.0,0.66) -- (7.0,1.72);
    \node[brakered, anchor=south] at (7.0,1.68) {$T_{\mathrm{safe}}$};
    \fill[thinkpurple!75] (0,1.17) rectangle (26.6,1.47);
    \node[white] at (16.5,1.32) {Qwen3-4B thinks for 256 tokens, about 26 s on the Orin};
    \draw[thinkpurple, line width=0.8pt] (26.6,1.1) -- (26.6,1.62);
    \node[thinkpurple, anchor=south] at (26.6,1.6) {answer};
    \fill[walkgreen] (0,0.51) rectangle (7.0,0.81);
    \node[white] at (3.5,0.66) {tracks $p_0$};
    \fill[brakered] (7.0,0.51) rectangle (8.4,0.81);
    \fill[gray!45] (8.4,0.51) rectangle (26.6,0.81);
    \node[black!75] at (17.5,0.66) {waits while the language model thinks on};
    \fill[checkgold] (26.6,0.51) rectangle (27.0,0.81);
    \fill[newteal] (27.0,0.51) rectangle (31.0,0.81);
    \node[newteal, anchor=south] at (29.4,0.83) {new plan};
    \node[brakered, anchor=north] at (7.7,0.47) {brakes};
    \node[checkgold!70!black, anchor=north] at (26.9,0.47) {timed check};
  \end{scope}
\end{tikzpicture}}
\caption{(a,b) A corridor sign and a delivery request. (c) The latest checked switch $T_{\mathrm{safe}}$ from tracking to braking. (d) The robot tracks its current plan during reasoning, brakes at the deadline, and adopts a revised plan after a timed clearance check.}
\label{fig:episode}
\end{figure}
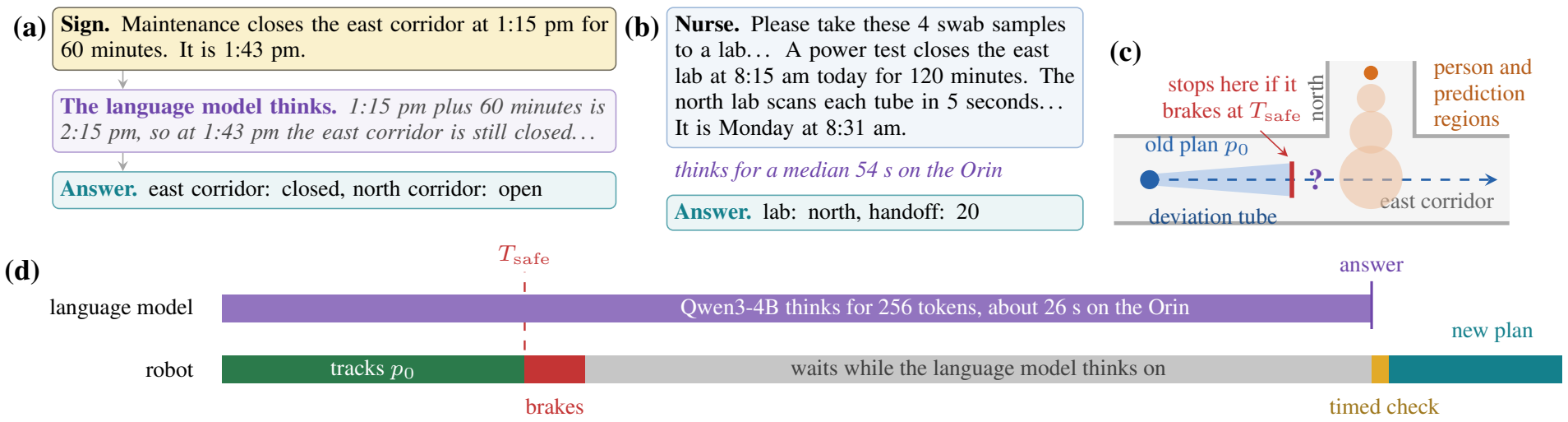

Mobile robots increasingly use natural-language information such as route restrictions and delivery requests to decide where to go and what to do (Figure~\ref{fig:episode}a,b).
Language models supply the open-world knowledge needed to interpret such messages~\cite{ahn2022saycan,huang2022inner}, yet accurate interpretation and capable motion control do not automatically compose into safe, timely missions~\cite{pecora2026missing}.
When Qwen3-4B~\cite{yang2025qwen3} answers immediately, it sends our robot into a closed corridor in about one run in four.
Full reasoning cuts this to about one run in a hundred, yet the answer can take most of a minute on a Jetson AGX Orin.
During that minute the people around the robot keep walking, so how long it may keep following a plan that the answer may overturn becomes a control decision in its own right.

Existing systems allocate reasoning effort~\cite{huang2026pace,liu2026rarrl,elboher2023cope,shperberg2021situated,sung2024effort}, accelerate inference~\cite{nguyen2026vlacpp,yang2026jetsonpi}, or overlap inference and execution~\cite{black2025rtc}.
In the closest work~\cite{pek2021failsafe}, a vehicle falls back on a verified emergency maneuver whenever a new plan fails a check that finishes within milliseconds and covers every position other road users could legally reach.
Onboard reasoning allows neither luxury as the language model's answer may arrive after most of a minute or never, and over such a wait the reach of a walking pedestrian spans the whole corridor.
We therefore bound pedestrians with calibrated prediction regions instead.

We present staleness deadlines for motion plans, which move the boundary between reasoning and control from the language model's answer to the robot's clock.
A plan grows stale as the people it was checked against move on, and the deadline marks the last moment at which the robot can still switch from the stale plan to braking without coming too close to anyone.
A clock monitor brakes at the deadline unless a plan built from the answer passes the same check in time (Figure~\ref{fig:episode}c,d).
Our contributions are as follows.
\begin{enumerate}[leftmargin=*,topsep=2pt,itemsep=1pt,parsep=0pt]
\item \textbf{Staleness deadlines} (Section~\ref{sec:method}) that carry fail-safe planning~\cite{kousik2020bridging,vaskov2019notatfault,pek2021failsafe} over to reasoning with no time bound.
The deadline plays the role of the time-to-react in~\cite{pek2021failsafe}, with conformal pedestrian regions and robot tubes in place of set-based reachability.
\item \textbf{Mission-level reasoning-budget qualification} that determines which prespecified fixed-budget and per-query policies keep close calls, wrong decisions and late arrivals below their targets with at least 95\% family-level confidence within each task and inference profile.
\item \textbf{Onboard evaluation} showing inexpensive offline replay anticipates losses of live Orin missions.
\end{enumerate}

\section{Problem setting}\label{sec:problem}
A planar robot with state $x\in\R^4$ (position, heading and speed) follows routes through a building while simulated pedestrians walk past.
In each mission, a language model interprets one message, either a corridor sign or a delivery request.
A graph planner turns that interpretation into a connected route, and a tracking controller drives the robot along it.

Each mission records three binary losses.
A \emph{close call} is contact with a wall or a pedestrian while the robot moves, or a separation below 10~cm.
A \emph{wrong decision} is entry into a closed corridor or a failed delivery.
A \emph{late arrival} exceeds the time limit $T_{\mathrm{late}}$ or never finishes.
Contact at rest counts separately as a \emph{stopped contact}.
A reasoning budget $k$ caps the tokens the language model may spend thinking before it answers.
A reasoning policy $q$ either fixes one budget for every query or uses a prespecified rule to choose a budget for each query.
The risk $R_i(q)$ is the probability of loss $i\in\{\mathrm{close},\mathrm{wrong},\mathrm{late}\}$ under that policy.
Let $\mathcal Q$ denote the family of reasoning policies fixed before certification.

While the language model thinks, a tracking controller $\pi_{p_0}$ follows the current plan $p_0$ until a switch time $s$, after which a braking controller $\pi_{\mathrm b}$ brings the robot to rest within $T_{\mathrm b}$.
We seek the latest safe switch $T_{\mathrm{safe}}$, together with a certified subset $\widehat{\mathcal Q}\subseteq\mathcal Q$ for each task and inference profile such that
\begin{equation}\label{eq:goal}
\mathbb P_{\mathrm{cal}}\big[R_i(q)\le\gamma_i\ \text{for every loss } i \text{ and every } q\in\widehat{\mathcal Q}\big]\ge 1-\xi,
\end{equation}
where $\gamma_i$ is the target risk of loss $i$, $\xi$ is the tolerated error of the selection, and $\mathbb P_{\mathrm{cal}}$ runs over missions drawn independently from a distribution $\mathcal D$.

\section{Method}\label{sec:method}
\paragraph{The robot's clock sets the deadline.}
We identify the robot's dynamics $\dot x=\Phi(x,u)\,\theta+d(x,u)$ from the transitions $\{(x_k,u_k,x_{k+1})\}_{k=1}^{20}$, with learned basis functions $\Phi$~\cite{iqbal2026zeroshot}, a parameter estimate $\hat\theta$ and a misfit $d$.
Rolling out a switch from tracking to braking at time $s$ gives a nominal trajectory $\bar x_s(\tau)$.
A deviation tube of radius $\Delta_s(\tau)$ around it bounds how far the executed trajectory may stray because of parameter error and misfit.
Each predicted pedestrian path carries a disc whose radius grows with the prediction horizon $\tau$~\cite{lindemann2023safe}.
The deadline is the latest candidate switch whose widened trajectory stays clear of walls and pedestrian discs.
\begin{equation}\label{eq:deadline}
T_{\mathrm{safe}}=\max\big\{s\in\mathcal S:\ h^{\mathrm w}(\bar x_s(\tau))-\Delta_s(\tau)\ge\varphi,\quad
h^{\mathrm p}_\tau(\bar x_s(\tau))-\Delta_s(\tau)\ge0\ \text{at checked moving }\tau\big\}.
\end{equation}
Here $\mathcal S=\{0,0.2,\ldots,10\}$~s holds the candidate switches, and $h^{\mathrm w}$ and $h^{\mathrm p}_\tau$ measure clearance to walls and pedestrians after body radii.
The wall floor $\varphi=\min(0,m_0)$ depends on the initial wall margin $m_0$, the widened wall clearance at the start of the check.
A robot that starts too close to a wall, with $m_0<0$, then only has to avoid getting closer, so it can still move away.
Following~\cite{vaskov2019notatfault}, the check skips steps at which the robot is fully at rest.
A clock monitor brakes at $T_{\mathrm{safe}}$ whether or not the language model has answered, and at once when no switch passes.

\paragraph{Calibration sizes both margins.}
Both margins come from simulated data through split conformal prediction.
For the pedestrian discs, we record how far each predicted path strays from the true one in $n_{\mathrm p}$ calibration scenes.
Each scene contributes its worst error, scaled by the growth of the disc over the horizon.
The disc radius is the quantile of these scores that leaves out a fraction $\delta$ of the scenes.
The tube follows the same recipe.
Its scores are the worst deviations of the robot during tracking and braking in $n_{\mathrm r}$ simulated deliberations, and its quantile leaves out a fraction $\zeta$.
A new segment drawn from the same distribution as the calibration data therefore stays inside both margins with probability at least $1-\delta-\zeta$, by a union bound over the two events.
Because the deadline alone triggers braking, this argument needs no bound on the language model's response time.

\paragraph{A new plan must pass the same check in time.}
The language model's answer marks corridors on the map as open or closed, and the planner rebuilds the route from the edited map.
The robot cannot switch to that route at once, because checking it takes time while the robot keeps moving.
The controller therefore schedules the switch for a fixed delay $\Delta_{\mathrm a}$ after the answer arrives.
It bounds where the robot may be at that moment by a box of states, calibrated on replay missions, and checks the new route from each corner of the box against the same walls, discs and tube.
The robot adopts the route only if this check finishes before the scheduled switch.
Otherwise the old deadline stays in force.
Whenever a check outlasts a control period, the controller repeats its last command.

\paragraph{Exact tests decide which reasoning policies pass.}
For each query, a per-query policy combines measured decision-error and response-time profiles with the time the robot can continue its current plan before it must brake, then chooses among the candidate budgets.
We run each policy on $N$ independent missions and count the number $n_i$ of missions that suffer loss $i$.
If the true risk of loss $i$ equals its target $\gamma_i$, the chance of observing $n_i$ or fewer such missions is $p_i=\mathbb P[\mathrm{Binomial}(N,\gamma_i)\le n_i]$, so a small $p_i$ is evidence that the risk lies below its target.
We use the conservative cutoff $\max_i p_i\le\xi/9$ for every reported policy.
For the seven policies in $\mathcal Q$, a policy that truly misses a target passes with probability at most $\xi/9$, so the chance that any such policy passes is at most $7\xi/9<\xi$.
This establishes the guarantee in Eq.~\eqref{eq:goal} for the prespecified family~\cite{angelopoulos2025ltt}.
The two fixed-budget follow-ups are exploratory and lie outside that family-level guarantee.

\section{Experiments}\label{sec:exp}
\paragraph{One building, two tasks, three inference profiles.}
The simulated building has 3~m corridors, two routes to a goal 25~m away, and four non-yielding pedestrians whose motion states are observed by the robot.
We record greedy FP32 Qwen3-4B answers on an RTX 5090 and replay them offline under three inference profiles, with 39 tokens/s on the desktop, 10.0 tokens/s on an idle Orin, and 5.9 tokens/s on an Orin under a 30~frame/s perception load.
Task time limits are 60/120~s for signs and 120/240~s for deliveries on the desktop/Orin.
A braking baseline stops while the language model reasons.
For each task and inference profile, we test four fixed budgets and three prespecified per-query rules, followed by exploratory fixed budgets of 384 and 640 tokens.
The rules already include both among their candidate budgets.
Each policy runs 1500 missions against targets of 3\% close calls, 5\% wrong decisions and 10\% late arrivals.

\paragraph{Reasoning fixes decisions, and the deadline saves time.}
Full reasoning reduces wrong decisions from 25\% to 1\% for signs and from 37\% to 2\% for deliveries.
The robot never collides with a wall or a pedestrian while it moves, and no policy records more than two close calls in 1500 missions.
Pedestrians who never yield still walk into the robot after it stops, in 23\% of the 63\,000 main-round missions.
On the desktop at 256 tokens, tracking until the deadline cuts late sign missions by 64\% against the braking baseline.
Tracking during deliberation thus turns waiting into progress.

\begin{figure}[t]
\centering
\includegraphics[width=\textwidth]{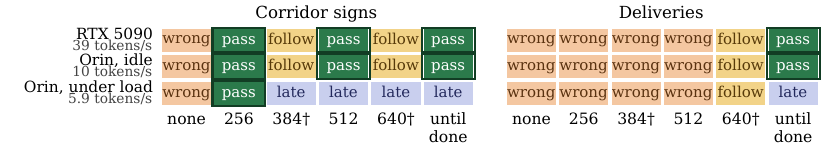}
\vspace{-2em}
\caption{Budget screening over 1500 missions per policy. ``Pass'' meets all targets, and ``follow'' with $^\dagger$ marks an exploratory fixed-budget follow-up.}
\label{fig:window}
\end{figure}

\begin{table}[b]
\centering
\footnotesize
\caption{Live / replay counts on the Orin, 150 missions per row. Loss agreement is computed only on missions where either run records a loss, and $\kappa$ is Cohen's kappa for stopped contacts.}
\label{tab:live}
\setlength{\tabcolsep}{3.5pt}
\begin{tabular}{llccccccc}
\toprule
Orin & task, budget & \begin{tabular}[c]{@{}c@{}}close\\calls\end{tabular} & \begin{tabular}[c]{@{}c@{}}wrong\\decisions\end{tabular} & \begin{tabular}[c]{@{}c@{}}late\\arrivals\end{tabular} & \begin{tabular}[c]{@{}c@{}}loss\\agreement\end{tabular} & \begin{tabular}[c]{@{}c@{}}stopped\\runs\end{tabular} & \begin{tabular}[c]{@{}c@{}}stopped\\$\kappa$\end{tabular} & \begin{tabular}[c]{@{}c@{}}thinking\\(s)\end{tabular}\\
\midrule
idle & signs, 256 & 0 / 0 & 9 / 7 & 0 / 1 & 7 / 10 & 59 / 49 & 0.68 & 26.4\\
idle & deliveries, 640 & 0 / 0 & 4 / 4 & 1 / 1 & 4 / 4 & 37 / 19 & 0.27 & 54.4\\
loaded & signs, 256 & 0 / 0 & 9 / 7 & 3 / 2 & 8 / 13 & 53 / 49 & 0.55 & 45.0\\
loaded & deliveries, 640 & 0 / 0 & 4 / 4 & 2 / 2 & 5 / 5 & 42 / 42 & 0.44 & 92.5\\
\bottomrule
\end{tabular}
\end{table}

\paragraph{Compute load narrows the viable reasoning budgets.}
Immediate answers make too many wrong decisions under every inference profile (Figure~\ref{fig:window}).
Full reasoning passes on the desktop and on the idle Orin, but under load it takes so long that too many missions arrive late.
Under load, only the 256-token budget survives for signs.
For deliveries, all three rules set in advance still pass the test, because they answer nearly every query at 640 tokens.

\paragraph{Replay anticipates onboard losses.}
In live missions, the Orin runs Qwen3-4B and controls a simulated robot in real time.
We run 150 fresh worlds per task and load condition, at 256 tokens for signs and 640 for deliveries, and then replay each world offline with its recorded answer (Table~\ref{tab:live}).
Most missions end without any loss, so we count agreement only where either run records one.
There, live and replay agree on all three losses in 24 of 32 missions, and all eight disagreements arise in signs, six from differing answers.
Stopped contacts occur in 32\% of live missions and agree less well.
On the Orin the timed check rejects two to seven times as many new plans, so the live robot stops more often.
The offline test therefore anticipates mission losses, while stopped contacts still need live measurement.

\bibliography{refs}

\end{document}